%% file: main.tex
\documentclass[10pt,twocolumn,letterpaper]{article}
\usepackage[accsupp]{axessibility}
\usepackage{cvpr}              

\usepackage{graphicx} 
\usepackage{rotating} 
\usepackage{tabularx} 
\usepackage{stfloats}
\usepackage{bibunits}

\input{preamble}

\definecolor{cvprblue}{rgb}{0.21,0.49,0.74}
\usepackage[pagebackref,breaklinks,colorlinks,citecolor=cvprblue]{hyperref}

\def\paperID{952} 
\def\confName{CVPR}
\def\confYear{2024}

\title{CARZero: Cross-Attention Alignment for Radiology Zero-Shot Classification}

\author{
    Haoran Lai$^{1,2,}$\thanks{These authors contributed equally to this work} \qquad 
    Qingsong Yao$^{3,}$\footnotemark[1] \qquad 
    Zihang Jiang$^{2,}$\thanks{Corresponding author} \qquad 
    Rongsheng Wang$^{2}$ \\
    Zhiyang He$^{4}$ \qquad 
    Xiaodong Tao$^{4}$ \qquad 
    S. Kevin Zhou$^{1,2,3,}$\footnotemark[2] \\
$^{1}$ School of Biomedical Engineering, Division of Life Sciences and Medicine, \\ 
University of Science and Technology of China, Hefei, Anhui, 230026, P.R.China \\
$^{2}$ Suzhou Institute for Advanced Research, University of Science and  Technology of  China, \\ Suzhou, Jiangsu, 215123, P.R.China \\
$^{3}$  Key Lab of Intelligent Information Processing of Chinese Academy of Sciences \\ (CAS), Institute of Computing Technology, CAS, Beijing 100190, China \\
$^{4}$ Medical Business Department, iFlytek Co.Ltd, Hefei 230088, China \\
{\tt\small \{haoranlai, rongsheng\_wang\}@mail.ustc.edu.cn, yaoqingsong19@mails.ucas.edu.cn} \\
{\tt\small \{zyh, xdtao\}@iflytek.com, jzh0103@ustc.edu.cn, s.kevin.zhou@gmail.com} }
\begin{document}
\maketitle
\input{sec/0_abstract}    
\input{sec/1_introduction}
\input{sec/2_related_work}
\input{sec/3_method}
\input{sec/4_experiments}

\input{sec/5_conclusion}
\input{sec/6_acknowledgment}

{
    \small
    \bibliographystyle{ieeenat_fullname}
    \bibliography{main}
}




\end{document}

%% file: preamble.tex
\usepackage[dvipsnames]{xcolor}

\newcommand{\algname} {CARZero\xspace}

\makeatletter

\DeclareRobustCommand\onedot{\futurelet\@let@token\@onedot}
\def\@onedot{\ifx\@let@token.\else.\nu11\fi\xspace}

\makeatother

%% file: sec/0_abstract.tex
\begin{abstract}
\vspace{-3mm}
The advancement of Zero-Shot Learning in the medical domain has been driven forward by using pre-trained models on large-scale image-text pairs, focusing on image-text alignment. However, existing methods primarily rely on cosine similarity for alignment, which may not fully capture the complex relationship between medical images and reports. To address this gap, we introduce a novel approach called Cross-Attention Alignment for Radiology Zero-Shot Classification (\algname). Our approach innovatively leverages cross-attention mechanisms to process image and report features, creating a Similarity Representation that more accurately reflects the intricate relationships in medical semantics. This representation is then linearly projected to form an image-text similarity matrix for cross-modality alignment. Additionally, recognizing the pivotal role of prompt selection in zero-shot learning, \algname incorporates a Large Language Model-based prompt alignment strategy. This strategy standardizes diverse diagnostic expressions into a unified format for both training and inference phases, overcoming the challenges of manual prompt design.  Our approach is simple yet effective, demonstrating state-of-the-art performance in zero-shot classification on five official chest radiograph diagnostic test sets, including remarkable results on datasets with long-tail distributions of rare diseases. This achievement is attributed to our new image-text alignment strategy, which effectively addresses the complex relationship between medical images and reports. Code and models are available at \url{https://github.com/laihaoran/CARZero}.
\end{abstract}

%% file: sec/1_introduction.tex
\begin{figure}[t]
  \centering
    \includegraphics[width=1.0\linewidth]{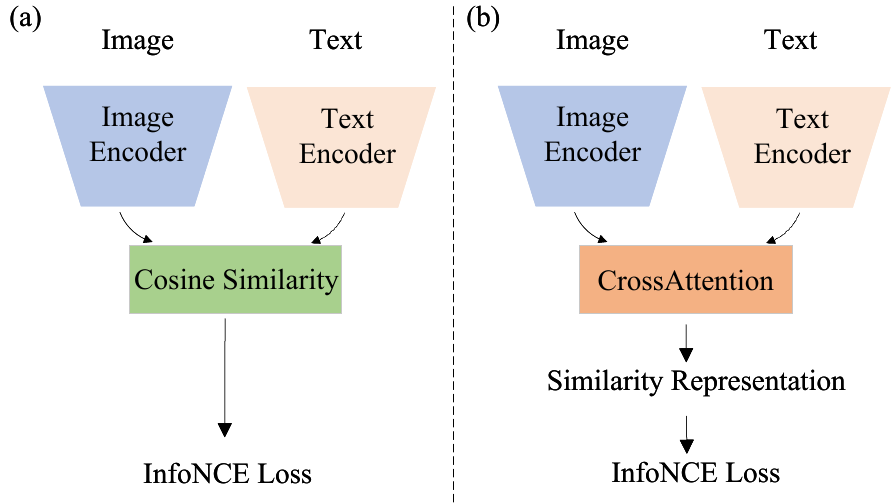}
    \caption{Comparison of the alignment scheme in Visual Language Pre-training: (left) handcrafted cosine similarity used in CLIP~\cite{conde2021clip} and CheXzero~\cite{tiu2022expert}; (right) our proposed cross-attention alignment leveraging a novel similarity representation.} \label{fig:compare}
    \vspace{-5mm}
\label{fig:teaser}
\end{figure}

\section{Introduction}
\label{sec:introduction}
Deep learning (DL) has achieved remarkable success in medical image recognition tasks. Prior studies~\cite{tran2021deep,chan2020computer,jamshidi2020artificial,lee2022deep} have harnessed DL techniques for diagnosing diseases, yielding impressive results. However, these efforts often rely on laborious and costly annotations from clinical experts. Additionally, to reach an acceptable accuracy level, model training requires an extensive collection of labeled data for each specific disease, which can be both challenging and time-consuming. To address these issues, recent works~\cite{boecking2022making, zhang2022contrastive,huang2021gloria,tiu2022expert,wu2023medklip,zhang2023knowledge} have utilized paired images and reports for cost-effective disease diagnosis through zero-shot learning (ZSL)~\cite{radford2021learning}. For rare diseases, where labeled training data is particularly scarce and difficult to obtain, ZSL emerges as a valuable tool for diagnosis.

Delving into the advanced medical ZSL methods, including ConVIRT~\cite{zhang2022contrastive}, GLoRIA~\cite{huang2021gloria}, CheXzero~\cite{tiu2022expert}, MedKLIP~\cite{wu2023medklip}, and KAD~\cite{zhang2023knowledge}, we discover that contrastive learning serves as a foundational approach, aiming at minimizing the cosine similarity between paired image-text samples and maximizing it between unpaired ones. However, compared to natural images and texts, the relationship between medical images and reports is significantly more complex. For example, radiologists tend to describe multiple findings, diseases, and their locations within a single report, drawing upon various visual clues present in the corresponding medical images. Hence, we hypothesize that relying solely on the hand-crafted cosine similarity to measure the complex relationships between medical reports and images might be suboptimal.

To address this issue, we propose a \algname method that leverages cross-attention alignment for radiology disease diagnosis within the setup of ZSL, a \textit{simple yet effective} method of learning a similarity measurement that robustly represents the similarity of the medical semantic context. Specifically, as shown in Fig.~\ref{fig:teaser}, we use cross-attention~\cite{vaswani2017attention} to compute global and local features from both modalities and result in features of mutual interaction, which is referred to the Similarity Representation (SimR). The SimR is then processed by a linear projector to obtain the final logit of similarity. Finally, we employ an InfoNCE loss~\cite{radford2021learning} to introduce comparisons between positive and negative pairs, optimizing the model to learn discriminative features.

Furthermore, considering that medical reports are quite specialized and complex, the task of human-designed prompts poses an additional challenge to medical zero-shot classification. Recently, there has been significant research efforts~\cite{zhou2022learning,tao2023galip,khattak2023maple,kawar2023imagic},  dedicated to optimizing prompts, mainly with strategies that adjust the adaptability of prompts for downstream tasks. In our \algname framework, we tackle this issue by aligning diagnostic prompts during both training and zero-shot inference phases. Fortunately, recent advancements in Large Language Models (LLMs)~\cite{zhao2023survey} have demonstrated significant capabilities in laguege comprehension and reformulation, enabling \algname to standardize the diverse expressions found in reports into a cohesive and unified prompt format. This not only mitigates the challenges of manual prompt design but also unifies the format of diagnosis to improve zero-shot inference performance.

Our \algname has been evaluated on a total of five public datasets. In particular, our \algname achieves the state-of-the-art (SOTA) AUC performance of 0.810 in  PadChest~\cite{bustos2020padchest}, which is a multi-label dataset with a long-tail distribution spanning 192 diseases. More surprisingly, our \algname achieves zero-shot performance scores of 0.811 on ChestXray14~\cite{wang2017chestx}, surpassing the SOTA performances of 0.794 achieved with fine-tuning on 1\% of the data.

To summarize, the contributions of this paper are listed as follows:
\begin{itemize}
  \item We propose a novel cross-attention alignment for medical images and reports, utilizing SimR to articulate the complex relationships between medical images and reports, effectively aligning the features of both vision and text domains.
  
  \item We employ LLM to reformulate medical reports into the unified prompt template, ensuring the alignment of diagnostic expression during both the training and zero-shot inference phases.

  \item Tremendous experiments on five large-scale radiology diagnosis datasets confirm the zero-shot capabilities of our \algname exceeding the SOTA zero-shot methods with a notable performance gap. Impressively, a significant improvement is achieved in diagnosing rare diseases. 
\end{itemize}

%% file: sec/2_related_work.tex
\section{Related Work}
\label{sec:related_work}

\subsection{Zero-shot classification}
For Vision Language Pretraining (VLP) tasks, previous works~\cite{li2019visualbert,li2021align} mainly use a fusion module to integrate image and text features, employing binary cross-entropy for classifying the combined features to determine if the image-text pair matches. Recently, CLIP~\cite{radford2021learning} introduced contrastive learning, which measures the cosine similarity between image and text features, aiming to maximize it between the matching image-text pairs and minimize the unpaired ones. This work significantly advances the development of VLP for ZSL in visual recognition tasks. Following CLIP, many studies~\cite{conde2021clip,ma2022ei,chen2023disco,dong2023maskclip,xie2023ra,varma2023villa,gandelsman2023interpreting,sun2023alpha} have utilized contrastive learning for aligning image-text, demonstrating the substantial potential of contrastive learning for VLP.

 In the medical domain, VLP has demonstrated remarkable performance in ZSL for disease diagnosis. There has been a succession of outstanding works~\cite{zhang2022contrastive,huang2021gloria,tiu2022expert,wu2023medklip,zhang2023knowledge}. ConVIRT~\cite{zhang2022contrastive} first introduces contrastive learning to align medical images and reports. CheXzero~\cite{tiu2022expert} leverages the CLIP trained by nature data as pre-trained weights to achieve commendable performance on medical data. GLoRIA~\cite{huang2021gloria} further introduces the integration of global and local features alignment.  MedKLIP~\cite{wu2023medklip} proposes the use of prior knowledge in the form of disease descriptions as an additional input to enhance representation learning. KAD~\cite{zhang2023knowledge} introduces word-based entity extraction to extract report information, thus improving the model's generalizability. For medical zero-shot classification tasks, these advancements represent meaningful progress. However, an important consideration they overlook is the intricate and nuanced relationship between medical images and reports, which is substantially more complex than the associations found in natural images and texts. Therefore, capturing the complex relationships between medical images and texts is key to improving the performance of medical zero-shot classification.

\subsection{Cross-Attention in Modality Alignment}
 Existing works of cross-attention in modality alignment can be divided into two types. The first type uses cross-attention for modality fusion, optimizing the fused features with an image-text matching loss (ITM). For example, ALBEF~\cite{li2021align} proposes a strategy of alignment before fusion, enabling the cross-attention module to merge the features of two modalities and optimize them using an ITM loss. BLIP~\cite{li2022blip} treats the image features as hidden states of text encoder to obtain fused modal features, which are then optimized with an ITM loss. The second type involves using cross-attention for modality projection transformation, followed by employing cosine similarity to calculate the similarity between the projected modalities and optimizing with an InfoNCE loss. For example, MGCA~\cite{wang2022multi} employs cross-attention to project local features of both modalities, aligning the projected features afterward by InfoNCE loss. TEFAL~\cite{ibrahimi2023audio} uses cross-attention to project video and audio on text, then optimizes these projections with an InfoNCE loss. 
 In contrast to these established methodologies, as shown in \Cref{fig:compare}, this paper introduces a novel third type, which employs cross-attention to directly generate a high-level Similarity Representation between two modalities, subsequently optimized using an InfoNCE loss. This high-level SimR effectively captures the complex relationship between the two modalities, especially for medical images and reports. Moreover, InfoNCE loss, compared to the ITM loss, uses more negative samples and a softmax that is highly comparative, along with a process of logit normalization, which is more suitable for modality alignment.

\subsection{Prompt Alignment}
In VLP models, the performance of zero-shot classification is highly dependent on the design of the prompt. Typically, human-designed prompts require extensive searching to find the optimal template, which is time-consuming and labor-intensive. To improve efficiency, CoOP~\cite{zhou2022learning} introduced the concept of automatic prompts by learning prompts for downstream tasks. MaPLe~\cite{khattak2023maple} proposes a multi-modal prompt fine-tuning strategy, emphasizing the interplay between text and images in prompt construction.
Imagic~\cite{kawar2023imagic} proposes a novel framework for text-guided image editing, enabling precise alterations that resonate with the image's semantic context. 
GALIP~\cite{tao2023galip} utilizes text-conditioned prompts to better adapt to downstream tasks, thereby improving complex image synthesis capabilities. Despite their effectiveness, these developments mainly leverage natural data, underscoring the significance of prompt design in the pre-training phase of cross-modality learning. Intuitively, using prompt templates as training data to align the training and testing prompts has been an effective method for prompt design. However, it is challenging to directly insert prompts into training data. Fortunately, Recent progress in LLMs~\cite{zhao2023survey} have shown enormous potential in semantic understanding, making it feasible to introduce prompt information into training texts. Therefore, leveraging large models for prompt alignment emerges as another key aspect in advancing medical zero-shot classification.

%% file: sec/3_method.tex
\section{Method}
\label{sec:method}

\begin{figure*}[t]
  \centering
    \includegraphics[width=0.95\linewidth]{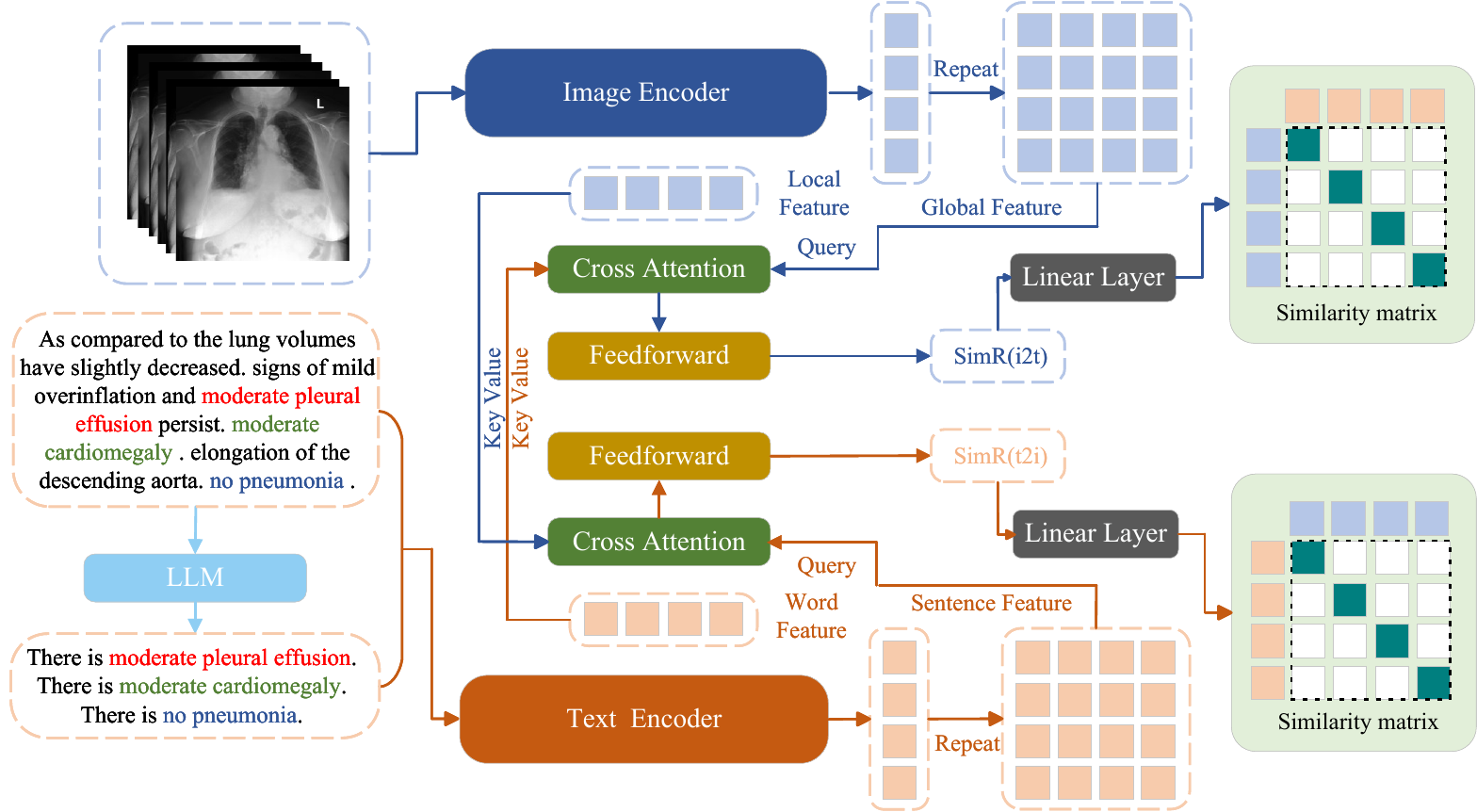}
    \caption{The \algname Network proposed in this paper consists of two stages. First, LLM is employed to generate prompt templates from medical reports. Second, text and vision encoders are used to extract features from image and text, which are fed into a cross-attention module to generate similarity for optimizing InfoNCE loss.} 
    \vspace{-1mm}
    \label{fig:Network}
    \vspace{-4mm}
\end{figure*}

In this section, we describe our proposed \algname framework for zero-shot classification. As illustrated in \Cref{fig:Network}, first, we introduce a Cross-Attention Alignment that generates a SimR to represent the relationship between images and reports. Then, a linear layer projects the SimR onto a similarity matrix, which is then optimized using the InfoNCE loss. Moreover, we propose an LLM-based Prompt Alignment method that integrates prompt templates into the training data.

\subsection{Feature Extraction}
Assume that the training dataset contains $N$ samples denoted as $D_{\text{train}} =  \left\{\left(x_{1}, y_{1}\right), \ldots, \left(x_{N}, y_{N}\right) \right\}$, where $x_{i} \in \mathbb{R}^{H \times W \times C}$ represents a CXR image and $y_{i}$  represents its corresponding medical report. $H$, $W$, and $C$ refer to height, width, and channels, respectively. As illustrated in \Cref{fig:Network}, we introduce individual components of our architecture for feature extraction of different modalities, including an image encoder $\Phi_{\text{image}}$ and a text encoder $\Phi_{\text{text}}$.

\noindent
{\bf Image Encoder } The image encoder is utilized to extract the image features at different levels, as shown in \cref{image_encoder}.
\begin{equation}
\label{image_encoder}[\boldsymbol{x}^{l}_{i},\boldsymbol{x}^{g}_{i}]  =  \Phi_{\text{image}}\left( x_{i} \right),
\end{equation}
where $\boldsymbol{x}^{l}_{i} \in \mathbb {R}^{L \times D}$ and $\boldsymbol{x}^{g}_{i} \in \mathbb {R}^{D}$ represent the local and global features, respectively. $L$ represents the number of local image patches and $D$ refers to the dimension of image features. The ViT-base~\cite{dosovitskiy2020image} model is adopted as the visual encoder in our experiments.

\noindent
 {\bf Text Encoder } As illustrated in \cref{text_encoder}, the text encoder is employed to extract text features from the given report. For each sample, a sentence is randomly selected from the report in each training iteration. In our experiments, to better adapt the text encoder for medical-related information extraction, we fine-tune BioBERT~\cite{lee2020biobert} with the clinical reports and use it for text encoding.
\begin{equation}
\label{text_encoder}
    [\boldsymbol{y}^{l}_{i}, \boldsymbol{y}^{g}_{i}] =  \Phi_{\text{text}}\left( y_{i} \right),
\end{equation}   
where $\boldsymbol{y}^{l}_{i} \in \mathbb {R}^{M \times D}$, $\boldsymbol{y}^{g}_{i} \in \mathbb {R}^{D}$ represent the word-based and sentence-based features, respectively. $M$ is the maximum text length and $D$ refers to the dimension of text features. In our experiment, the dimensions of final output features from both the image and text encoders are the same.

\subsection{Cross-Attention Alignment}
Due to the complex relationship between medical images and reports, a cross-attention alignment module is proposed to align the feature spaces between images and reports. The objective is to employ cross-attention to obtain the SimR, which serves as a high-level representation of the correlation between the images and reports. During the training phase, the number of images and texts in a batch is equal, denoted as $I$ for the number of images and $T$ for the number of texts. For text-to-image (`t2i') alignment, sentence-based features $\boldsymbol{y}^{g} \in \mathbb {R}^{T \times D}$ are used as the query. To calculate SimR, the dimension of $\boldsymbol{y}^{g}$ is expanded and repeated to match the number of images, resulting in $\hat{\boldsymbol{y}}^{g} \in \mathbb {R}^{T \times I \times D}$. Then, the local features of images, $\boldsymbol{x}^{l} \in \mathbb {R}^{ L \times I \times D}$ are used as key and value. Finally, the output from the cross-attention module is SimR, $SR_{\text{t2i}} \in \mathbb {R}^{T \times I \times D}$. This high-dimensional, learned representation is considered as a rich descriptor of the similarity between texts and images, effectively capturing their complex relationship. 
\begin{equation}
 [Q, K, V] = [W^{Q} \hat{\boldsymbol{y}}^{g}, W^{K} \boldsymbol{x}^{l},  W^{V} \boldsymbol{x}^{l}];
\end{equation}
\begin{equation}
    \text{CrossAtt}(Q, K, V) = \text{Softmax}\left(\frac{QK^T}{\sqrt{d_k}}\right)V ;
\end{equation}
\begin{equation}
    SR_{\text{t2i}} = \text{Feedforward} \left( \text{CrossAtt}(Q, K, V) \right).
\end{equation}
\noindent where $W^{Q}$, $W^{K}$, and $W^{V}$ are the weights for linear projection. $d_{k}$ is the feature dimension of $K$, which is equal to $D$.  Following this, a linear projection is applied to map this high-dimensional representation to a low-dimensional space to derive the similarity matrix. 
\begin{equation}
    S_{\text{t2i}} = \text{Linear}(SR_{\text{t2i}})\in \mathbb {R}^{T \times I};
\end{equation}
\noindent Optimization is performed using the InfoNCE loss.
 \begin{equation}
\label{simple_contrastive_loss_t2i}
\mathcal{L}_{\text{t2i}}=-\log \frac{e^{ S^{i,i}_{\text{t2i}} }}{\sum_{k=1}^I e^{S^{i,k}_{\text{t2i}}}} -\log \frac{e^{ S^{i,i}_{\text{t2i}} }}{\sum_{k=1}^T e^{S^{k,i}_{\text{t2i}}}};
\end{equation}
\noindent $S^{i,i}_{\text{t2i}}$ refers to a specific element in the similarity matrix, representing the similarity score between the $i$th image-text pair in the batch. Similarly, $S^{i,k}_{\text{t2i}}$ indicates the similarity score between the $i$th text and the $k$th image in the batch.

Similarly, for image-to-text (`i2t') alignment, global features from images are used as the query, and the word-based features of texts serve as the key and value. This process also yields a high-dimensional SimR, $SR_{\text{i2t}} \in \mathbb {R}^{I \times T \times D}$, for image-to-text, which is then projected using a linear layer to obtain the image-to-text similarity matrix for computation of InfoNCE loss. 
 \begin{equation}
\label{simple_contrastive_loss_i2t}
\mathcal{L}_{\text{i2t}}=-\log \frac{e^{ S^{i,i}_{\text{i2t}} }}{\sum_{k=1}^T e^{S^{i,k}_{\text{i2t}}}} -\log \frac{e^{ S^{i,i}_{\text{i2t}} }}{\sum_{k=1}^I e^{S^{k,i}_{\text{i2t}}}};
\end{equation}
The final objective function for the \algname model is the summation of $\mathcal{L}_{\text{t2i}}$ and $\mathcal{L}_{\text{i2t}}$, given by:
\begin{equation}
\mathcal{L}= \mathcal{L}_{\text{t2i}} + \mathcal{L}_{\text{i2t}}.
\end{equation}


\subsection{LLM-based Prompt Alignment}
To align the prompts used during training and inference phases, we incorporate prompt templates into the training data using LLMs. Prompting instruction is utilized to generate the prompt template within the training data, with details shown in the supplementary. By leveraging the LLM’s exceptional capability for semantic understanding, fixed prompt templates are introduced into the training data. The templates generated by the LLM are then merged with the original reports to create enhanced reports for training. During the inference phase, the prompt template, ``There is [disease].'', is employed for zero-shot classification. This strategy leverages the LLM's advanced capabilities in semantic comprehension to ensure that the training data is enriched with consistent and relevant prompt templates, thereby facilitating a more effective and aligned application during both training and inference stages.


%% file: sec/4_experiments.tex
\section{Experiments}
\label{sec:experiments}

\subsection{Dataset}
\noindent
{\bf MIMIC-CXR~\cite{johnson2019mimic} } In our experiments, we conducted model pretraining using the MIMIC-CXR dataset, a publicly available collection of chest radiographs paired with radiology text reports. The MIMIC-CXR dataset comprises 377,110 images corresponding to 227,835 radiographic studies conducted on 65,379 patients. Each radiographic study is accompanied by a radiology report and the corresponding chest X-ray image, which may be in either frontal or lateral views. The radiology report serves as a comprehensive summary provided by radiologists, encompassing various sections such as examination, indication, impression, findings, technique, and comparison. In our methodology, we selectively retain only the findings and impressions sections from these reports. Moreover, only frontal views of CXRs are used for \algname training.

\noindent
{\bf Open-I~\cite{demner2016preparing} } Open-I contains 3,851 reports and 7,470 Chest X-ray images, which includes manual annotations for 18 different multi-label diseases. We evaluate the \algname for zero-shot classification on Open-I.

\noindent
{\bf PadChest~\cite{bustos2020padchest} } PadChest has 160,868 chest X-ray images labeled with 192 different diseases, which is a long-tailed distribution dataset. 39,053 (27\%) samples are manually annotated by board-certified radiologists. For evaluation purposes, we only test on samples annotated by board-certified radiologists. Additionally, we select categories with fewer than 10 samples, totaling 20 classes, designated as \textbf{PadChest20}, to evaluate the performance of \algname on rare diseases.

\noindent
{\bf ChestXray14~\cite{wang2017chestx} }
NIH ChestXray14 has 112,120 chest X-ray images with 14 disease labels from
30,805 unique patients. The official test set released by the NIH, comprising 22,433 images, are distinctively annotated by board-certified radiologists. For evaluation purposes, we only test on the official test set.

\noindent
{\bf CheXpert~\cite{irvin2019chexpert} }
CheXpert has 224,316 CXRs collected from 65,240 patients. The official test set contains 500 patients annotated by a consensus of 5 board-certified radiologists~\cite{rajpurkar2021chexternal}. We evaluate on 5 observations:  Atelectasis, Cardiomegaly, Consolidation, Edema, and Pleural Effusion.

\noindent
{\bf ChestXDet10~\cite{liu2020chestx} }
ChestX-Det10 is a subset of NIH ChestXray14, which is consisting of 3543
CXRs with box-level annotations provided by 3 board-certified radiologists of 10
diseases. The official test set contains 542 CXRs with 10 diseases and corresponding box-level annotations. We evaluate the zero-shot classification and grounding ability in the official test set.

\subsection{Evaluation Metric}
For the multi-label test dataset, we adopt Area under the ROC Curve (AUC), Matthews Correlation Coefficient (MCC), F1 score (F1), and Accuracy (ACC) as metrics for evaluating zero-shot classification tasks. For assessing zero-shot grounding tasks, we specifically utilize the  Pointing Game~\cite{zhang2018top} metric.

\begin{table*}
    \centering
    \setlength{\tabcolsep}{ 11pt}
   \begin{tabular}{@{}lcccccc@{}}
    \toprule
    Method & Open-I & PadChest & PadChest20 & ChestXray14 & CheXpert & ChestXDet10  \\
    \midrule
    MedCLIP~\cite{wang2022medclip} & 0.551 & 0.508 & 0.501 & 0.564 & 0.744 & 0.571 \\
    BiomedCLIP~\cite{zhang2023large} & 0.577 & 0.513 & 0.510 & 0.639 & 0.677 & 0.630 \\
    GLoRIA~\cite{huang2021gloria} & 0.589 & 0.565 & 0.558 & 0.610 & 0.750 & 0.645 \\
    BioViL~\cite{bannur2023learning} & 0.702 & 0.655 & 0.608 & 0.729 & 0.789 &  0.708 \\
    CheXzero~\cite{tiu2022expert} & 0.726 & 0.648 & 0.644  & 0.712 & 0.889 & 0.640 \\
    MedKLIP~\cite{wu2023medklip} & 0.759  & 0.629 & 0.688  & 0.726 & 0.879 & 0.713 \\
    KAD~\cite{zhang2023knowledge} & 0.807 & 0.750 &  0.735 & 0.789 & 0.905 & 0.735 \\
    \algname & \textbf{0.838} & \textbf{0.810} & \textbf{0.837}  &\textbf{0.811} & \textbf{0.923} & \textbf{0.796} \\
    \bottomrule
    \end{tabular}
    \vspace{-2mm}
    \caption{Comparative analysis of existing zero-shot classification approaches on five official multi-label CXR datasets evaluated by AUC.}
    \label{tab:competition}
    \vspace{-3mm}
\end{table*}

\subsection{Implementation Details}
In our experiments, ViT-B/16 is used as the image encoder, which utilizes M3AE~\cite{chen2022multi} for pretraining on the MIMIC dataset. For the text encoder, BioBERT is fine-tuned using texts from both MIMIC and PadChest datasets.  Given that PadChest reports are in Spanish, they are translated into English. The LLM named Spark\footnote{\url{https://xinghuo.xfyun.cn/}} is employed to intelligently insert prompt templates into the training dataset.

The cross-attention module shares weights for `i2t' and `t2i' alignments. The images are resized to a uniform shape of $224 \times 224$. Commonly used data augmentation including random horizontal flips, random affine transformations, and color jittering are adopted. For each report, we divide it into multiple sentences with a maximum length of 97, and one of the sentences is randomly selected at each training iteration. The optimizer used is Adam with a learning rate set to 5e-5. The code is based on the PyTorch framework. All experiments are conducted with an A800 GPU.

During the inference phase, when prompt alignment strategy is applied, we set the prompt template as ``There is [disease]'', denoted as $P_1$. Without prompt alignment, an empirically optimized prompt ``A disease of [disease]'', denoted as $P_2$, is found to yield the best performance.

\begin{figure*}[t]
  \centering
    \includegraphics[width=1.0\linewidth]{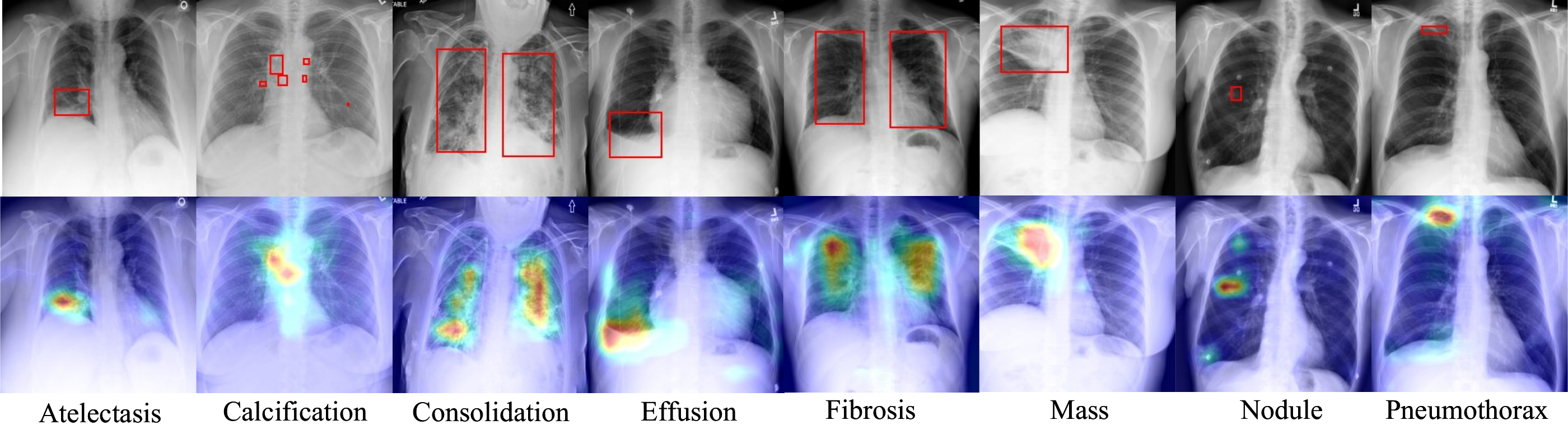}
      \vspace{-6mm}
    \caption{Visualization of attention map in \algname on ChestXDet10. The red boxes indicate the corresponding ground truth of detection. Highlighted pixels represent higher activation weights correlating specific words with regions in the image.} \label{fig:Visual}
      \vspace{-2mm}
\end{figure*}


\begin{table}
\vspace{-2mm}
  \centering
    \setlength{\tabcolsep}{ 20pt}
  \begin{tabular}{@{}lc@{}}
    \toprule
    Method & ChestXray14 \\
    \midrule
    GLoRIA~\cite{huang2021gloria} & 0.707 \\
    MedKLIP~\cite{wu2023medklip} & 0.772 \\
    MGCA~\cite{wang2022multi} & 0.782 \\
    KAD~\cite{zhang2023knowledge} & 0.787 \\
    MRM~\cite{zhou2023advancing} & 0.794 \\
    \midrule
    \algname & \textbf{0.811}\\
    \bottomrule
  \end{tabular}
  \vspace{-2mm}
  \caption{Comparison of the performance of existing methods fine-tuned on 1\% data versus \algname in zero-shot classification on ChestXray14 by AUC.}
  \label{tab:finetuning}
  \vspace{-5mm}
\end{table}

\begin{table*}
  \centering
  \begin{tabular}{@{}lccccccccccc@{}}
    \toprule
    Method & Mean & ATE & CALC & CONS & EFF & EMPH & FIB & FX & MASS & NOD & PTX\\
    \midrule
    GLoRIA~\cite{huang2021gloria} & 0.367 & 0.479 & 0.053 & 0.737 & 0.528 & 0.667 & 0.366 & 0.013 & 0.533 & 0.156 & 0.143 \\
    MedKLIP~\cite{wu2023medklip} & 0.481 & 0.625 & 0.132 & \textbf{0.837} & 0.675 & 0.734 & 0.305 & \textbf{0.224} & \textbf{0.733} & 0.312 & 0.229 \\
    KAD~\cite{zhang2023knowledge} & 0.391 & \textbf{0.646} & 0.132 & 0.699 & 0.618 & 0.644 & 0.244 & 0.199 & 0.267 & \textbf{0.316} & 0.143 \\
    \algname & \textbf{0.543} & 0.604 & \textbf{0.184} & 0.824 & \textbf{0.782} & \textbf{0.846} & \textbf{0.561} & 0.184 & 0.700 & 0.286 & \textbf{0.457} \\
    \bottomrule
  \end{tabular}
  \vspace{-2mm}
  \caption{Comparison of various methods on the ChestXDet10 dataset for zero-shot grounding using the pointing game. The abbreviations ATE, CALC, CONS, EFF, EMPH, FIB, FX, MASS, NOD, and PTX correspond to Atelectasis, Calcification, Consolidation, Effusion, Emphysema, Fibrosis, Fracture, Mass, Nodule, and Pneumothorax, respectively.}
  \label{tab:grounding}
  \vspace{-5mm}
\end{table*}

\subsection{Comparison with State-of-the-art Methods}
 As shown in \Cref{tab:competition}, we compare the performance of existing SOTA methods in CXR zero-shot classification on five officially released test sets. All test sets are manually annotated to ensure reliability. To ensure a fair comparison, all methods use their respective published models for inference. Among these, GLoRIA~\cite{huang2021gloria}, MedKLIP~\cite{wu2023medklip}, KAD~\cite{zhang2023knowledge}, and our \algname all utilize both global and local feature information for alignment. Ultimately, our proposed \algname achieves better performance on the AUC metric across all test sets. Particularly for the PadChest dataset, which includes a long-tail distribution of 192 diseases, our method achieves an AUC of 0.810, indicating its good generalization performance on long-tail distribution data. Specifically, we achieve exceptional performance in rare disease zero-shot classification on PadChest20, attributable to our effective image-text alignment method. This demonstrates the significant potential of our approach for diagnosing rare diseases. Moreover, we compare zero-shot classification performance of our method with the results of existing works using 1\% labeled data for fine-tuning. As shown in \Cref{tab:finetuning}, our zero-shot classification performance even surpasses that of the existing methods fine-tuned on 1\%  data, demonstrating the strength of our approach.
 This might be due to cross-attention alignment used in \algname generates high-level SimR to represent the relationships between medical images and reports, effectively measuring their complex relationships. Furthermore, the process of projecting SimR from high to low dimension using a learnable projection matrix fully exploits the associative information in SimR, thereby effectively aligning the feature spaces of images and texts. Additionally, our proposed prompt alignment strategy aligns prompts during the training and inference phases, eliminating the need to search for prompts, thereby enhancing the generalizability of \algname to the prompts and improving zero-shot classification performance.

Additionally, we test the performance of \algname in zero-shot grounding. We use the attention map from the cross-attention for grounding prediction. The aim of zero-shot grounding is to match prompt tokens with image tokens. As shown in \Cref{tab:grounding}, our \algname achieves the best performance. This is likely because our method directly uses SimR, generated by cross-attention alignment, to represent the similarity between images and texts. Therefore, the attention map in cross-attention effectively reflects the association between images and texts. The outstanding performance of our \algname demonstrates that our method can effectively align the feature spaces of images and reports, and successfully match the lesion areas in the images with the prompts.

\subsection{Visualization}
As shown in \Cref{fig:Visual}, we present the visualization results of \algname. We perform linear interpolation on the attention map from the cross-attention alignment to obtain a pseudocolor image of the same size as the original image. At the same time, we display the names of the lesions and their corresponding locations. From \Cref{fig:Visual}, it is evident that \algname effectively captures the correlation between disease-related words and the corresponding lesion areas in the images, providing strong interpretability for our method. For small lesions, our method can also precisely detect them from the images. For example, as shown in \Cref{fig:Visual}, for pneumothorax, \algname accurately locates the corresponding lesion areas and localizes the full lesion. This may be attributed to the attention mechanism. In this paper, we propose a cross-attention alignment strategy, using the attention mechanism to directly align images and texts. This method can effectively capture text-related information from images based on a given prompt, which is demonstrated by \Cref{tab:grounding} and \Cref{fig:Visual}.

\subsection{Ablation Study}
{\bf Ablation Study of Modules } To validate the effectiveness of the prompt alignment and cross-attention alignment proposed in this paper, we design experiments using the CLIP framework as the baseline, incorporating each of these modules separately. In scenarios without prompt alignment, we compare the performance of using different prompts $P_1$ and $P_2$. As shown in \Cref{tab:ablation_module} (a vs. c and b vs. c), the prompt alignment strategy aligns the \algname with $P_1$, achieving performance comparable to the empirically tuned one. This improvement is attributed to the model's alignment during the training phase with prompt $P_1$, leading to stronger generalization on $P_1$.

The comparison in \Cref{tab:ablation_module} (a vs. d and b vs. e) demonstrates that introducing cross-attention alignment further enhances model performance. While CLIP utilizes global features from images and texts for alignment using cosine similarity, our method employs both local and global features for alignment. Additionally, high-level SimR is used to decipher the feature associations between images and texts, effectively aligning them and achieving superior performance. \Cref{tab:ablation_module} (f) highlights the effectiveness of combining prompt alignment and cross-attention alignment for zero-shot classification. The integrated use of these techniques not only consolidates the strengths of each individual component but also helps the model to effectively generalize to diverse zero-shot scenarios.

\begin{table}
    \centering
    \begin{tabular*}{\columnwidth}{@{\extracolsep{\fill}}c|ccc|cccc}
        \toprule
         \# & PT & PA & CA & AUC & MCC & F1 & ACC  \\
        \midrule
        a & $P_1$ &  &  & 0.764 & 0.230 & 0.247 & 0.792   \\
        b & $P_2$ &  &  & 0.796 & 0.255 & 0.271 & 0.821   \\
        c & $P_1$ & \checkmark & & 0.795 & \textbf{0.288} & \textbf{0.290} & 0.866   \\
        d & $P_1$ & & \checkmark & 0.781 & 0.217 & 0.230 & 0.816   \\
        e & $P_2$ & & \checkmark & 0.801 & 0.241 & 0.253 & 0.821   \\
        f & $P_1$ & \checkmark & \checkmark & \textbf{0.810} & 0.257 & 0.270 & \textbf{0.867} \\
        \bottomrule
    \end{tabular*}
    \vspace{-2mm}
     \caption{Results on ChestXray14 for the ablation study of various modules. Here, 'PT' denotes the prompt template used in the inference stage, 'PA' represents prompt alignment, and 'CA' stands for cross-attention alignment.} 
    \label{tab:ablation_module}
    \vspace{-5mm}
\end{table}

\noindent
{\bf Ablation Study of Feature} In the context of cross-attention alignment, the query must utilize global features, while the key and value have three options: global only, local only, or a combination of both global and local features. We experiment with all three choices, as demonstrated in \Cref{tab:ablation_global_local}. The results show that using only global features yields the worst performance, while combining global and local features achieves optimal performance. The performance of using only local features is close to that of combining both global and local features. First, global features are an aggregation of local features, and this aggregation process may lead to the loss of detailed information. Hence, relying solely on global features is insufficient. Compared to global features, local features encompass richer semantic information, including details and positional information, thereby ensuring the completeness of feature information. Moreover, since global features are the aggregation of local features and match the global features of the query, integrating both global and local features can achieve the best performance, which aligns with intuition.

\begin{table}
    \centering
    \begin{tabular*}{\columnwidth}{@{\extracolsep{\fill}}cc|cccc}
        \toprule
         Global & Local & AUC & MCC & F1 & ACC  \\
        \midrule
        \checkmark & & 0.799 & 0.248 & 0.264 & 0.856   \\
         & \checkmark & 0.810 & 0.257 & 0.270 & 0.867   \\
         \checkmark & \checkmark & \textbf{0.810} & \textbf{0.259} & \textbf{0.276} & \textbf{0.880} \\
        \bottomrule
    \end{tabular*}
    \vspace{-2mm}
     \caption{Results on ChestXray14 for key and value choice in cross-attention alignment.} 
    \label{tab:ablation_global_local}
    \vspace{-2mm}
\end{table}

\begin{table}
    \centering
    \begin{tabular*}{\columnwidth}{@{\extracolsep{\fill}}lccccc}
        \toprule
        Method & AUC & MCC & F1 & ACC  \\
        \midrule
        $cos(x_{y}, y_{x})$ & 0.804 & 0.225 & 0.228 & 0.813   \\
         $cos(x_{y}, y) + cos(y_{x}, x)$ & 0.805 & 0.111 & 0.130 & 0.420  \\
        $\text{Linear}(SR)$ & 0.810 & 0.259 & 0.276 & \textbf{0.880}   \\
        $\text{MLP}(SR)$ & \textbf{0.811} & \textbf{0.265} & \textbf{0.284} & 0.878 \\
        \bottomrule
    \end{tabular*}
    \vspace{-2mm}
     \caption{Results on ChestXray14 for Processing of SimR. Here, $cos\left( \cdot\right)$ denotes cosine similarity, $x_{y}$ refers to $SR_{\text{i2t}}$ as the projection of image features onto text, and $y_{x}$ indicates $SR_{\text{t2i}}$ as the projection of text features onto image.}
    \label{tab:ablation_simr}
    \vspace{-4mm}
\end{table}

\subsection{Processing SimR}
To further explore the role of SimR in image-text alignment and validate the rationale behind using SimR in our study, we conduct the following investigations: (1) \textbf{Cosine similarity computation.} We consider SimR as a modality projection transformation. Following this, we can calculate the alignment of two modalities using cosine similarity, similar to the methods described in~\cite{ibrahimi2023audio}. Consequently, we design two sets of experiments: one directly computes the cosine similarity between two projected features, while the other computes the cosine similarity between each projected feature and the original global features of images and texts, respectively. (2) \textbf{Direct projection to low-dimension similarity.} We continue to treat SimR as a high-level associative representation between images and texts. To more effectively extract the relationships between images and reports from SimR, we replace the simple linear layer with a multi-layer perceptron (MLP).

As shown in \Cref{tab:ablation_simr}, cosine similarity computation proves less effective than direct projection to low-dimension similarity. This may be caused by the fact that the features obtained through cross-attention alignment, as high-level SimR, can effectively represent the relationship between images and medical reports, eliminating the need for further similarity calculations, which is the core of our method. Moreover, cosine similarity, a non-parametric similarity metric, is less effective compared to our learnable similarity strategy, which better captures the complex relationships between images and medical reports, thus achieving superior performance. This indicates that using cosine similarity alone may not adequately measure the complex relationships between medical reports and images. Lastly, compared to simple linear projection, MLP employs more complex non-linear projection to reduce SimR from high to low dimensions. Experimental results demonstrate that MLP outperforms simple linear projection. This suggests that a simple linear layer might not be sufficient to fully extract the relationships between images and texts from SimR, which contains complex image-text relationships. Therefore, a complex MLP better leverages SimR's advantages.

%% file: sec/5_conclusion.tex
\section{Conclusion, Limitation and Impact}
\label{sec:conclusion}
This paper proposes \algname that achieves high-performing zero-shot classification. First, we propose a novel cross-attention alignment strategy, innovatively using the features generated by cross-attention as the SimR between images and reports. Subsequently, we use a linear layer to project SimR into a low-dimension similarity for aligning images and reports. Extensive experiments prove that the information contained in this SimR is highly effective for image-text alignment. Moreover, we propose a novel LLM-based prompt alignment strategy, integrating prompt templates into the training data to achieve training and inference prompt alignment. Finally, \algname achieves SOTA performance on five publicly released official test sets, demonstrating its effectiveness.

\noindent
\textbf{Limitations and Future Work } First, our method mainly focuses on zero-shot classification. 
The \algname framework employs a cross-attention module as the core mechanism for aligning images and texts,
which can be directly used as a classifier, thus offering certain advantages in fine-tuning. Therefore, future work could involve experiments in fine-tuning tasks. Secondly, our method primarily focuses on the complex relationship between medical images and reports, introducing a cross-attention alignment strategy to address it. We believe this method also has excellent generalizability to natural data. Hence, we plan to experiment with natural data in the future to further validate the effectiveness of the cross-attention alignment strategy.

\noindent
\textbf{Impact} The \algname proposed in this paper achieves state-of-the-art performance in CXR zero-shot classification tasks, including excellent performance on the long-tail dataset, PadChest, which includes 192 categories. This is significant for the diagnosis of rare diseases. In addition, ZSL greatly reduces the manpower cost of radiology experts and allows for low-cost acquisition of more training data, which is crucial for advancing chest AI diagnostics.

%% file: sec/6_acknowledgment.tex
\section{Acknowledgment}
\label{sec:acknowledgment}
This work was supported by Natural Science Foundation of China under Grant 62271465 and Open Fund Project of Guangdong Academy of Medical Sciences, China (No. YKY-KF202206).

%% file: main.bbl
\begin{thebibliography}{45}
\providecommand{\natexlab}[1]{#1}
\providecommand{\url}[1]{\texttt{#1}}
\expandafter\ifx\csname urlstyle\endcsname\relax
  \providecommand{\doi}[1]{doi: #1}\else
  \providecommand{\doi}{doi: \begingroup \urlstyle{rm}\Url}\fi

\bibitem[Boecking et~al.(2022)Boecking, Usuyama, Bannur, Castro, Schwaighofer,
  Hyland, Wetscherek, Naumann, Nori, Alvarez-Valle, et~al.]{boecking2022making}
Benedikt Boecking, Naoto Usuyama, Shruthi Bannur, Daniel~C Castro, Anton
  Schwaighofer, Stephanie Hyland, Maria Wetscherek, Tristan Naumann, Aditya
  Nori, Javier Alvarez-Valle, et~al.
\newblock Making the most of text semantics to improve biomedical
  vision--language processing.
\newblock In \emph{European conference on computer vision}, pages 1--21.
  Springer, 2022.

\bibitem[Bustos et~al.(2020)Bustos, Pertusa, Salinas, and De~La
  Iglesia-Vaya]{bustos2020padchest}
Aurelia Bustos, Antonio Pertusa, Jose-Maria Salinas, and Maria De~La
  Iglesia-Vaya.
\newblock Padchest: A large chest x-ray image dataset with multi-label
  annotated reports.
\newblock \emph{Medical image analysis}, 66:\penalty0 101797, 2020.

\bibitem[Chan et~al.(2020)Chan, Hadjiiski, and Samala]{chan2020computer}
Heang-Ping Chan, Lubomir~M Hadjiiski, and Ravi~K Samala.
\newblock Computer-aided diagnosis in the era of deep learning.
\newblock \emph{Medical physics}, 47\penalty0 (5):\penalty0 e218--e227, 2020.

\bibitem[Chen et~al.(2023)Chen, Qi, Wang, and Zhang]{chen2023disco}
Yihao Chen, Xianbiao Qi, Jianan Wang, and Lei Zhang.
\newblock Disco-clip: A distributed contrastive loss for memory efficient clip
  training.
\newblock In \emph{Proceedings of the IEEE/CVF Conference on Computer Vision
  and Pattern Recognition}, pages 22648--22657, 2023.

\bibitem[Chen et~al.(2022)Chen, Du, Hu, Liu, Li, Wan, and Chang]{chen2022multi}
Zhihong Chen, Yuhao Du, Jinpeng Hu, Yang Liu, Guanbin Li, Xiang Wan, and
  Tsung-Hui Chang.
\newblock Multi-modal masked autoencoders for medical vision-and-language
  pre-training.
\newblock In \emph{International Conference on Medical Image Computing and
  Computer-Assisted Intervention}, pages 679--689. Springer, 2022.

\bibitem[Conde and Turgutlu(2021)]{conde2021clip}
Marcos~V Conde and Kerem Turgutlu.
\newblock Clip-art: Contrastive pre-training for fine-grained art
  classification.
\newblock In \emph{Proceedings of the IEEE/CVF Conference on Computer Vision
  and Pattern Recognition}, pages 3956--3960, 2021.

\bibitem[Demner-Fushman et~al.(2016)Demner-Fushman, Kohli, Rosenman, Shooshan,
  Rodriguez, Antani, Thoma, and McDonald]{demner2016preparing}
Dina Demner-Fushman, Marc~D Kohli, Marc~B Rosenman, Sonya~E Shooshan, Laritza
  Rodriguez, Sameer Antani, George~R Thoma, and Clement~J McDonald.
\newblock Preparing a collection of radiology examinations for distribution and
  retrieval.
\newblock \emph{Journal of the American Medical Informatics Association},
  23\penalty0 (2):\penalty0 304--310, 2016.

\bibitem[Dong et~al.(2023)Dong, Bao, Zheng, Zhang, Chen, Yang, Zeng, Zhang,
  Yuan, Chen, et~al.]{dong2023maskclip}
Xiaoyi Dong, Jianmin Bao, Yinglin Zheng, Ting Zhang, Dongdong Chen, Hao Yang,
  Ming Zeng, Weiming Zhang, Lu Yuan, Dong Chen, et~al.
\newblock Maskclip: Masked self-distillation advances contrastive
  language-image pretraining.
\newblock In \emph{Proceedings of the IEEE/CVF Conference on Computer Vision
  and Pattern Recognition}, pages 10995--11005, 2023.

\bibitem[Dosovitskiy et~al.(2020)Dosovitskiy, Beyer, Kolesnikov, Weissenborn,
  Zhai, Unterthiner, Dehghani, Minderer, Heigold, Gelly,
  et~al.]{dosovitskiy2020image}
Alexey Dosovitskiy, Lucas Beyer, Alexander Kolesnikov, Dirk Weissenborn,
  Xiaohua Zhai, Thomas Unterthiner, Mostafa Dehghani, Matthias Minderer, Georg
  Heigold, Sylvain Gelly, et~al.
\newblock An image is worth 16x16 words: Transformers for image recognition at
  scale.
\newblock \emph{arXiv preprint arXiv:2010.11929}, 2020.

\bibitem[et~al.(2023{\natexlab{a}})]{bannur2023learning}
Bannur et al.
\newblock Learning to exploit temporal structure for biomedical vision-language
  processing.
\newblock 2023{\natexlab{a}}.

\bibitem[et~al.(2023{\natexlab{b}})]{varma2023villa}
Varma et al.
\newblock Villa: Fine-grained vision-language representation learning from
  real-world data.
\newblock 2023{\natexlab{b}}.

\bibitem[et~al.(2022)]{wang2022medclip}
Wang et al.
\newblock Medclip: Contrastive learning from unpaired medical images and text.
\newblock \emph{EMNLP}, 2022.

\bibitem[et~al.(2023{\natexlab{c}})]{zhang2023large}
Zhang et al.
\newblock Large-scale domain-specific pretraining for biomedical
  vision-language processing.
\newblock 2023{\natexlab{c}}.

\bibitem[Gandelsman et~al.(2023)Gandelsman, Efros, and
  Steinhardt]{gandelsman2023interpreting}
Yossi Gandelsman, Alexei~A Efros, and Jacob Steinhardt.
\newblock Interpreting clip's image representation via text-based
  decomposition.
\newblock \emph{arXiv preprint arXiv:2310.05916}, 2023.

\bibitem[Huang et~al.(2021)Huang, Shen, Lungren, and Yeung]{huang2021gloria}
Shih-Cheng Huang, Liyue Shen, Matthew~P Lungren, and Serena Yeung.
\newblock Gloria: A multimodal global-local representation learning framework
  for label-efficient medical image recognition.
\newblock In \emph{Proceedings of the IEEE/CVF International Conference on
  Computer Vision}, pages 3942--3951, 2021.

\bibitem[Ibrahimi et~al.(2023)Ibrahimi, Sun, Wang, Garg, Sanan, and
  Omar]{ibrahimi2023audio}
Sarah Ibrahimi, Xiaohang Sun, Pichao Wang, Amanmeet Garg, Ashutosh Sanan, and
  Mohamed Omar.
\newblock Audio-enhanced text-to-video retrieval using text-conditioned feature
  alignment.
\newblock In \emph{Proceedings of the IEEE/CVF International Conference on
  Computer Vision}, pages 12054--12064, 2023.

\bibitem[Irvin et~al.(2019)Irvin, Rajpurkar, Ko, Yu, Ciurea-Ilcus, Chute,
  Marklund, Haghgoo, Ball, Shpanskaya, et~al.]{irvin2019chexpert}
Jeremy Irvin, Pranav Rajpurkar, Michael Ko, Yifan Yu, Silviana Ciurea-Ilcus,
  Chris Chute, Henrik Marklund, Behzad Haghgoo, Robyn Ball, Katie Shpanskaya,
  et~al.
\newblock Chexpert: A large chest radiograph dataset with uncertainty labels
  and expert comparison.
\newblock In \emph{Proceedings of the AAAI conference on artificial
  intelligence}, pages 590--597, 2019.

\bibitem[Jamshidi et~al.(2020)Jamshidi, Lalbakhsh, Talla, Peroutka, Hadjilooei,
  Lalbakhsh, Jamshidi, La~Spada, Mirmozafari, Dehghani,
  et~al.]{jamshidi2020artificial}
Mohammad Jamshidi, Ali Lalbakhsh, Jakub Talla, Zden{\v{e}}k Peroutka, Farimah
  Hadjilooei, Pedram Lalbakhsh, Morteza Jamshidi, Luigi La~Spada, Mirhamed
  Mirmozafari, Mojgan Dehghani, et~al.
\newblock Artificial intelligence and covid-19: deep learning approaches for
  diagnosis and treatment.
\newblock \emph{Ieee Access}, 8:\penalty0 109581--109595, 2020.

\bibitem[Johnson et~al.(2019)Johnson, Pollard, Berkowitz, Greenbaum, Lungren,
  Deng, Mark, and Horng]{johnson2019mimic}
Alistair~EW Johnson, Tom~J Pollard, Seth~J Berkowitz, Nathaniel~R Greenbaum,
  Matthew~P Lungren, Chih-ying Deng, Roger~G Mark, and Steven Horng.
\newblock Mimic-cxr, a de-identified publicly available database of chest
  radiographs with free-text reports.
\newblock \emph{Scientific data}, 6\penalty0 (1):\penalty0 317, 2019.

\bibitem[Kawar et~al.(2023)Kawar, Zada, Lang, Tov, Chang, Dekel, Mosseri, and
  Irani]{kawar2023imagic}
Bahjat Kawar, Shiran Zada, Oran Lang, Omer Tov, Huiwen Chang, Tali Dekel, Inbar
  Mosseri, and Michal Irani.
\newblock Imagic: Text-based real image editing with diffusion models.
\newblock In \emph{Proceedings of the IEEE/CVF Conference on Computer Vision
  and Pattern Recognition}, pages 6007--6017, 2023.

\bibitem[Khattak et~al.(2023)Khattak, Rasheed, Maaz, Khan, and
  Khan]{khattak2023maple}
Muhammad~Uzair Khattak, Hanoona Rasheed, Muhammad Maaz, Salman Khan, and
  Fahad~Shahbaz Khan.
\newblock Maple: Multi-modal prompt learning.
\newblock In \emph{Proceedings of the IEEE/CVF Conference on Computer Vision
  and Pattern Recognition}, pages 19113--19122, 2023.

\bibitem[Lee et~al.(2020)Lee, Yoon, Kim, Kim, Kim, So, and
  Kang]{lee2020biobert}
Jinhyuk Lee, Wonjin Yoon, Sungdong Kim, Donghyeon Kim, Sunkyu Kim, Chan~Ho So,
  and Jaewoo Kang.
\newblock Biobert: a pre-trained biomedical language representation model for
  biomedical text mining.
\newblock \emph{Bioinformatics}, 36\penalty0 (4):\penalty0 1234--1240, 2020.

\bibitem[Lee et~al.(2022)Lee, Liu, Kim, Chen, Sun, Rogers, Chung, and
  Weng]{lee2022deep}
Junghwan Lee, Cong Liu, Junyoung Kim, Zhehuan Chen, Yingcheng Sun, James~R
  Rogers, Wendy~K Chung, and Chunhua Weng.
\newblock Deep learning for rare disease: A scoping review.
\newblock \emph{Journal of Biomedical Informatics}, page 104227, 2022.

\bibitem[Li et~al.(2021)Li, Selvaraju, Gotmare, Joty, Xiong, and
  Hoi]{li2021align}
Junnan Li, Ramprasaath Selvaraju, Akhilesh Gotmare, Shafiq Joty, Caiming Xiong,
  and Steven Chu~Hong Hoi.
\newblock Align before fuse: Vision and language representation learning with
  momentum distillation.
\newblock \emph{Advances in neural information processing systems},
  34:\penalty0 9694--9705, 2021.

\bibitem[Li et~al.(2022)Li, Li, Xiong, and Hoi]{li2022blip}
Junnan Li, Dongxu Li, Caiming Xiong, and Steven Hoi.
\newblock Blip: Bootstrapping language-image pre-training for unified
  vision-language understanding and generation.
\newblock In \emph{International Conference on Machine Learning}, pages
  12888--12900. PMLR, 2022.

\bibitem[Li et~al.(2019)Li, Yatskar, Yin, Hsieh, and Chang]{li2019visualbert}
Liunian~Harold Li, Mark Yatskar, Da Yin, Cho-Jui Hsieh, and Kai-Wei Chang.
\newblock Visualbert: A simple and performant baseline for vision and language.
\newblock \emph{arXiv preprint arXiv:1908.03557}, 2019.

\bibitem[Liu et~al.(2020)Liu, Lian, and Yu]{liu2020chestx}
Jingyu Liu, Jie Lian, and Yizhou Yu.
\newblock Chestx-det10: chest x-ray dataset on detection of thoracic
  abnormalities.
\newblock \emph{arXiv preprint arXiv:2006.10550}, 2020.

\bibitem[Ma et~al.(2022)Ma, Zhao, Lin, Kale, Wang, Yu, Gu, Choudhary, and
  Xie]{ma2022ei}
Haoyu Ma, Handong Zhao, Zhe Lin, Ajinkya Kale, Zhangyang Wang, Tong Yu,
  Jiuxiang Gu, Sunav Choudhary, and Xiaohui Xie.
\newblock Ei-clip: Entity-aware interventional contrastive learning for
  e-commerce cross-modal retrieval.
\newblock In \emph{Proceedings of the IEEE/CVF Conference on Computer Vision
  and Pattern Recognition}, pages 18051--18061, 2022.

\bibitem[Radford et~al.(2021)Radford, Kim, Hallacy, Ramesh, Goh, Agarwal,
  Sastry, Askell, Mishkin, Clark, et~al.]{radford2021learning}
Alec Radford, Jong~Wook Kim, Chris Hallacy, Aditya Ramesh, Gabriel Goh,
  Sandhini Agarwal, Girish Sastry, Amanda Askell, Pamela Mishkin, Jack Clark,
  et~al.
\newblock Learning transferable visual models from natural language
  supervision.
\newblock In \emph{International conference on machine learning}, pages
  8748--8763. PMLR, 2021.

\bibitem[Rajpurkar et~al.(2021)Rajpurkar, Joshi, Pareek, Ng, and
  Lungren]{rajpurkar2021chexternal}
Pranav Rajpurkar, Anirudh Joshi, Anuj Pareek, Andrew~Y Ng, and Matthew~P
  Lungren.
\newblock Chexternal: Generalization of deep learning models for chest x-ray
  interpretation to photos of chest x-rays and external clinical settings.
\newblock In \emph{Proceedings of the Conference on Health, Inference, and
  Learning}, pages 125--132, 2021.

\bibitem[Sun et~al.(2023)Sun, Fang, Wu, Zhang, Zang, Kong, Xiong, Lin, and
  Wang]{sun2023alpha}
Zeyi Sun, Ye Fang, Tong Wu, Pan Zhang, Yuhang Zang, Shu Kong, Yuanjun Xiong,
  Dahua Lin, and Jiaqi Wang.
\newblock Alpha-clip: A clip model focusing on wherever you want.
\newblock \emph{arXiv preprint arXiv:2312.03818}, 2023.

\bibitem[Tao et~al.(2023)Tao, Bao, Tang, and Xu]{tao2023galip}
Ming Tao, Bing-Kun Bao, Hao Tang, and Changsheng Xu.
\newblock Galip: Generative adversarial clips for text-to-image synthesis.
\newblock In \emph{Proceedings of the IEEE/CVF Conference on Computer Vision
  and Pattern Recognition}, pages 14214--14223, 2023.

\bibitem[Tiu et~al.(2022)Tiu, Talius, Patel, Langlotz, Ng, and
  Rajpurkar]{tiu2022expert}
Ekin Tiu, Ellie Talius, Pujan Patel, Curtis~P Langlotz, Andrew~Y Ng, and Pranav
  Rajpurkar.
\newblock Expert-level detection of pathologies from unannotated chest x-ray
  images via self-supervised learning.
\newblock \emph{Nature Biomedical Engineering}, 6\penalty0 (12):\penalty0
  1399--1406, 2022.

\bibitem[Tran et~al.(2021)Tran, Kondrashova, Bradley, Williams, Pearson, and
  Waddell]{tran2021deep}
Khoa~A Tran, Olga Kondrashova, Andrew Bradley, Elizabeth~D Williams, John~V
  Pearson, and Nicola Waddell.
\newblock Deep learning in cancer diagnosis, prognosis and treatment selection.
\newblock \emph{Genome Medicine}, 13\penalty0 (1):\penalty0 1--17, 2021.

\bibitem[Vaswani et~al.(2017)Vaswani, Shazeer, Parmar, Uszkoreit, Jones, Gomez,
  Kaiser, and Polosukhin]{vaswani2017attention}
Ashish Vaswani, Noam Shazeer, Niki Parmar, Jakob Uszkoreit, Llion Jones,
  Aidan~N Gomez, {\L}ukasz Kaiser, and Illia Polosukhin.
\newblock Attention is all you need.
\newblock \emph{Advances in neural information processing systems}, 30, 2017.

\bibitem[Wang et~al.(2022)Wang, Zhou, Wang, Vardhanabhuti, and
  Yu]{wang2022multi}
Fuying Wang, Yuyin Zhou, Shujun Wang, Varut Vardhanabhuti, and Lequan Yu.
\newblock Multi-granularity cross-modal alignment for generalized medical
  visual representation learning.
\newblock \emph{Advances in Neural Information Processing Systems},
  35:\penalty0 33536--33549, 2022.

\bibitem[Wang et~al.(2017)Wang, Peng, Lu, Lu, Bagheri, and
  Summers]{wang2017chestx}
Xiaosong Wang, Yifan Peng, Le Lu, Zhiyong Lu, Mohammadhadi Bagheri, and
  Ronald~M Summers.
\newblock Chestx-ray8: Hospital-scale chest x-ray database and benchmarks on
  weakly-supervised classification and localization of common thorax diseases.
\newblock In \emph{Proceedings of the IEEE conference on computer vision and
  pattern recognition}, pages 2097--2106, 2017.

\bibitem[Wu et~al.(2023)Wu, Zhang, Zhang, Wang, and Xie]{wu2023medklip}
Chaoyi Wu, Xiaoman Zhang, Ya Zhang, Yanfeng Wang, and Weidi Xie.
\newblock Medklip: Medical knowledge enhanced language-image pre-training.
\newblock \emph{medRxiv}, pages 2023--01, 2023.

\bibitem[Xie et~al.(2023)Xie, Sun, Xiong, Zheng, Zhao, and Zhou]{xie2023ra}
Chen-Wei Xie, Siyang Sun, Xiong Xiong, Yun Zheng, Deli Zhao, and Jingren Zhou.
\newblock Ra-clip: Retrieval augmented contrastive language-image pre-training.
\newblock In \emph{Proceedings of the IEEE/CVF Conference on Computer Vision
  and Pattern Recognition}, pages 19265--19274, 2023.

\bibitem[Zhang et~al.(2018)Zhang, Bargal, Lin, Brandt, Shen, and
  Sclaroff]{zhang2018top}
Jianming Zhang, Sarah~Adel Bargal, Zhe Lin, Jonathan Brandt, Xiaohui Shen, and
  Stan Sclaroff.
\newblock Top-down neural attention by excitation backprop.
\newblock \emph{International Journal of Computer Vision}, 126\penalty0
  (10):\penalty0 1084--1102, 2018.

\bibitem[Zhang et~al.(2023)Zhang, Wu, Zhang, Xie, and Wang]{zhang2023knowledge}
Xiaoman Zhang, Chaoyi Wu, Ya Zhang, Weidi Xie, and Yanfeng Wang.
\newblock Knowledge-enhanced visual-language pre-training on chest radiology
  images.
\newblock \emph{Nature Communications}, 14\penalty0 (1):\penalty0 4542, 2023.

\bibitem[Zhang et~al.(2022)Zhang, Jiang, Miura, Manning, and
  Langlotz]{zhang2022contrastive}
Yuhao Zhang, Hang Jiang, Yasuhide Miura, Christopher~D Manning, and Curtis~P
  Langlotz.
\newblock Contrastive learning of medical visual representations from paired
  images and text.
\newblock In \emph{Machine Learning for Healthcare Conference}, pages 2--25.
  PMLR, 2022.

\bibitem[Zhao et~al.(2023)Zhao, Zhou, Li, Tang, Wang, Hou, Min, Zhang, Zhang,
  Dong, et~al.]{zhao2023survey}
Wayne~Xin Zhao, Kun Zhou, Junyi Li, Tianyi Tang, Xiaolei Wang, Yupeng Hou,
  Yingqian Min, Beichen Zhang, Junjie Zhang, Zican Dong, et~al.
\newblock A survey of large language models.
\newblock \emph{arXiv preprint arXiv:2303.18223}, 2023.

\bibitem[Zhou et~al.(2023)Zhou, Lian, Wang, and Yu]{zhou2023advancing}
Hong-Yu Zhou, Chenyu Lian, Liansheng Wang, and Yizhou Yu.
\newblock Advancing radiograph representation learning with masked record
  modeling.
\newblock \emph{arXiv preprint arXiv:2301.13155}, 2023.

\bibitem[Zhou et~al.(2022)Zhou, Yang, Loy, and Liu]{zhou2022learning}
Kaiyang Zhou, Jingkang Yang, Chen~Change Loy, and Ziwei Liu.
\newblock Learning to prompt for vision-language models.
\newblock \emph{International Journal of Computer Vision}, 130\penalty0
  (9):\penalty0 2337--2348, 2022.

\end{thebibliography}
